\documentclass{article}

\usepackage{arxiv}

\usepackage{tikz}
\usetikzlibrary{fit}
\usetikzlibrary{positioning}
\usepackage{listings}
\usepackage[utf8]{inputenc} 
\usepackage[T1]{fontenc}    
\usepackage{url}            
\usepackage{booktabs}       
\usepackage{amsfonts}       
\usepackage{microtype}      
\usepackage{lipsum}			
\usepackage{graphicx}
\usepackage[backend=biber]{biblatex}
\newenvironment{tightcenter}{%
	\setlength\parskip{0pt}
	\begin{center}
	}{%
	\end{center}
}
\usepackage{soul}
\usepackage{pgfplots}
\usepackage{mdframed}
\usepackage{color}
\usepackage{neuralnetwork}
\usepackage{pythonhighlight}
\PassOptionsToPackage{table}{xcolor}
\usepackage{colortbl}
\usepackage{caption} 
\definecolor{mygreen}{RGB}{153, 247, 160}
\definecolor{myred}{RGB}{245, 137, 180}
\definecolor{myblue}{RGB}{152, 178, 250}
\definecolor{myyellow}{RGB}{252, 240, 71}
\DeclareRobustCommand{\hlgreen}[1]{{\sethlcolor{mygreen}\hl{#1}}}

\DeclareRobustCommand{\hlblue}[1]{{\sethlcolor{myblue}\hl{#1}}}

\title{Automatic Generation of Word Problems for Academic Education via Natural Language Processing (NLP)}

\author{\hspace{1mm}Stanley U.~Keller\\
	Department of Industrial Engineering\\
	DHBW Center for Advanced Studies\\
	Heilbronn \\
	\href{mailto:stanley.keller122@icloud.com}{\texttt{stanley.keller122@icloud.com}} \\
}

\renewcommand{\headeright}{}
\renewcommand{\undertitle}{}
\renewcommand{\shorttitle}{Automatic Generation of Word Problems for Academic Education via Natural Language Processing (NLP)}

\usepackage[
colorlinks=true,
urlcolor=blue,
linkcolor=black,
citecolor=.
]{hyperref}
\hypersetup{
pdftitle={Automatic Generation of Word Problems for Academic Education},
pdfauthor={Stanley \^{U}.~Keller},
pdfkeywords={Word Problem Generation, Mathematical Statistics, NLP},
}

\begin{document}
\maketitle

\begin{abstract}
	Digital learning platforms enable students to learn on a flexible and individual schedule as well as providing instant feedback mechanisms. The field of STEM education requires students to solve numerous training exercises to grasp underlying concepts. It is apparent that there are restrictions in current online education in terms of exercise diversity and individuality. Many exercises show little variance in structure and content, hindering the adoption of abstraction capabilities by students. 
	This thesis proposes an approach to generate diverse, context rich word problems.  In addition to requiring the generated language to be grammatically correct, the nature of word problems implies additional constraints on the validity of contents. 
	The proposed approach is proven to be effective in generating valid word problems for mathematical statistics. The experimental results present a tradeoff between generation time and exercise validity. The system can easily be parametrized to handle this tradeoff according to the requirements of specific use cases. 
\end{abstract}

\keywords{Word Problem Generation \and Mathematical Statistics \and NLP}

\section{Introduction}
\subsection{Motivation for on-demand Education Resources}

The era of online education comes with new opportunities in terms of student engagement. Digital learning platforms enable students to learn on a flexible and individual schedule as well as providing instant and thorough feedback mechanisms (cf. \cite{cavalcanti2021automatic}, p. 7). Individual student errors can be highlighted in solution paths and taken into account for future training exercises. One area where the aforementioned aspects are present is regarding training exercises for STEM education. Many topics in that field require students to solve numerous training exercises in order to grasp and abstract underlying concepts (cf. (\cite{adams2020shaping}, p. 122), (\cite{wiggins}, p. 4)). It is apparent that there are restrictions in classical offline education in terms of teacher and exercise availability, which prevents the fulfillment of truly individual education, both in terms of quantity and individuality of training data.  

The resulting demand for a practically unlimited amount of training data currently leads to limitation of exercise diversity. As training exercises need to be verified for validity in order to be useful, the verification process, which almost always implies human engagement, imposes a bottleneck for exercise generation.  A common remedy has been to reduce verification efforts by limiting exercise complexity.  This is achieved via populating a pre-validated exercise-skeleton with varying values, i.e. exchanging numbers (cf. (\cite{nabla}), (\cite{education})). This tradeoff, in many cases, entails repetitive exercises.

Repetitive training exercises lead to inferior real-world applicability of concepts (cf. \cite{garfield1995students}, pp. 28-32). This symptom is borne in the observation that students, once they have identified a structured approach to solving an exercise, will preferably apply the same approach to a new problem instance. If the learned approach is rather shallow, which is the case if students merely remember which numbers to plug into which formula based on number location in an exercise, it cannot be applied to real-world situations that most likely do not adhere to the same learned structure (cf. \cite{garfield1995students}, pp. 28-32). Exchanging numbers in a rigid exercise-skeleton leads to the adoption of such a shallow structure (cf. (\cite{vincent2008mathematics}, pp. 102-103), (\cite{petocz2007materials}, pp. 166-167)).

Diverse training exercises and participation in exercise generation, on the other hand, enforce the adoption of a thorough understanding of underlying concepts (cf. "active involvement" in \cite{garfield1995students}, p. 30). In order to enforce abstraction the constituent structure of the exercise should vary instead of merely exchanging numbers and contextual entities. To allow for the generation of diverse exercises, the verification bottleneck needs to be alleviated or avoided altogether. 

\subsection{Contributions of this Thesis} \label{Contributions of this thesis}

This thesis is guided by the research question of how diverse word problems in mathematical statistics can be synthesized in an automated fashion. In addition to requiring the generated language to be grammatically correct, the nature of word problems implies additional constraints on the validity of contents. 

For the scope of this thesis a word problem is defined as an ordered set $E=(C, Q)$, consisting of an exercise context $C$ and respective questions $Q$. In contrast to basic language modelling, word problems underly the constraint that, in addition to containing fluent and coherent text, they must be valid. An exercise is defined to be valid if it meets the following requirements:
\begin{enumerate}
	\item \textit{Completeness} \\
	The context must contain all information required to answer all questions
	\item \textit{Correctness} \\
	Information contained in the context must satisfy mathematical assumptions implied by the exercise type
	\item \textit{Consistency} \\
	Information contained in the context must not contain any contradictions
	\item \textit{Unequivocacy} \\
	All questions and all context information must be unambiguous
\end{enumerate}
Examples for each requirement are listed in table \ref{tab:requirements-examples}.

\begin{table}[]
	\centering
	\resizebox{\textwidth}{!}{%
		\begin{tabular}{|l|l|}
			\hline
			\rowcolor[HTML]{000000} 
			\multicolumn{1}{|c|}{\cellcolor[HTML]{000000}{\color[HTML]{FFFFFF} \textbf{Example}}} &
			\multicolumn{1}{c|}{\cellcolor[HTML]{000000}{\color[HTML]{FFFFFF} \textbf{Issue}}} \\ \hline
			\begin{tabular}[c]{@{}l@{}}Human height can be assumed to be normally distributed with a mean of 1.8m.\\ What is the chance that a person's height is measured at above 1.9m?\end{tabular} &
			\begin{tabular}[c]{@{}l@{}}Missing information about the standard deviation\\ $\rightarrow$ \textcolor{red}{not complete}\end{tabular} \\ \hline
			A traffic light is red with a chance of -10\% or green with a chance of 110\%. &
			\begin{tabular}[c]{@{}l@{}}Information does not adhere to mathematical assumptions\\ $\rightarrow$ \textcolor{red}{not correct}\end{tabular} \\ \hline
			A traffic light is either red with a chance of 60\% or green with a chance of 65\%. &
			\begin{tabular}[c]{@{}l@{}}Conflicting information\\ $\rightarrow$  \textcolor{red}{not consistent}\end{tabular} \\ \hline
			\begin{tabular}[c]{@{}l@{}}Dogs live 8 years on average, whereas cats live 10 years.\\ Waht is the average lifespan of an animal?\end{tabular} &
			\begin{tabular}[c]{@{}l@{}}Unclear task\\ $\rightarrow$  \textcolor{red}{not unequivocal}\end{tabular} \\ \hline
		\end{tabular}%
	}
	\caption{Examples of requirements posed on word problems in mathematical statistics (own illustration)}
	\label{tab:requirements-examples}
\end{table}

The desire for diversity among generated exercises originates from the intention to prevent memorization of surface structure and instead encourage abstraction capabilities of students. To enforce abstraction the constituent structure of the exercise should vary instead of merely exchanging numerical values and contextual entities in a fixed structure. Similarly, exercises should not only respect grammatical rules, but also maintain semantic coherence across sentences. Simply embedding relevant information into a general context without any coherence would again lead to memorization of surface structure, since relevant information could easily be distinguished by irrelevant information.

Consequently, a system that is tasked to adhere to all aforementioned requirements has to ensure both text fluency and diversity as well as guarantee content validity. This in turn requires managing the tradeoff between methods associated to those requirements, namely Neural Language Models (NLM) and template-based techniques. NLMs have empirically proven their ability to generate human-like, fluent text (cf. \cite{brown2020language}, p. 28), though they are hard to control with respect to validity constraints (cf. \cite{DBLP:journals/corr/abs-2005-00683}, p. 1). Template-based techniques on the other hand allow the developer complete control over generated phrases but, being predefined by nature, lack diversity.

\newpage

Based on the introduction above the following research objectives are derived:

\textbf{Research Question}\\
How can diverse word problems in mathematical statistics be synthesized in an automated fashion?

\textbf{Research Objectives}
\begin{enumerate}
	\item Develop a novel approach to synthesize natural language subject to general content validity constraints
	\item Apply the novel approach to the task of word problem generation for academic education in mathematical statistics
	\item Demonstrate the effectiveness of the novel approach in an experimental setting
\end{enumerate}

The remainder of this thesis is structured as follows. Chapter 2 presents related work on word problem generation. Chapter 3 details the proposed method for word problem generation and highlights key design choices of the approach. Chapter 4 presents the results of experiments conducted in the light of research objective 3. The remaining chapters are used to discuss the approach ex-post and conclude the thesis.

The source code for this thesis is available at \url{https://github.com/Stansuro/ExGen}.\\
An interactive online demo of the code is available at \href{https://colab.research.google.com/drive/1TXZViybIiwRhj_7usPyfWWOIuRmsxnza?usp=sharing}{Google Colab}.

\section{Related Work}

In (\cite{zhou2019towards}, pp. 494-503) the authors aim to generate word problems by encoding both the target equation and the desired context into a single embedding and decoding it into the tokens of the word problem. While they employ simple lookup embeddings for the desired context, the equation decoder is based on recurrent neural networks.

A similar approach is chosen by (\cite{liu2021mathematical}, pp. 3-6). The authors deviate from (\cite{zhou2019towards}) by encoding the target equation via a Graph Neural Network and incorporating a VAE architecture into the generation process. This approach allows them to sample from a distribution of problem embeddings at decoding time, enabling a great deal of exercise diversity. Additionally, they incorporate a commonsense knowledge graph that is also embedded by a Graph Neural Network to introduce even more diversity. While the generated exercises show strong diversity, the authors note that their model in some cases has trouble generating valid exercises in that it sometimes breaches the constraints for completeness and language fluency as documented in table \ref{tab:error-example}. 

\begin{table}[!ht]
	\footnotesize
	\begin{center}
		\begin{tabular}{l|p{4.5cm}lp{0.5cm}}
			\toprule
			Missing information & Teacher Mr.Huang and his 35 students come to row the boat. They find 6 more small boats than big boats. There are 10 more students in the small boat than in the big boat. How many big boats are there? \\
			\midrule
			Language disfluency & There are two types of heads: cockroaches and ants, cockroaches have 1 more head than ants, and cockroaches have \textcolor{red}{0 more than ant legs}, how many cockroaches and ants respectively? \\
			\bottomrule
		\end{tabular}
	\end{center}
	\caption{Typical problems of word problem generation (cf. \cite{liu2021mathematical}, p. 9)}
	\label{tab:error-example}
\end{table}

The authors of (\cite{liyanage2019multilanguage}, p. 3) also identify the task as constrained text generation and define the following constraints:
\begin{enumerate}
	\item \textit{Mathematical correctness} \\
	In their case referring to the size relation between two values or algebraic satisfiability
	\item \textit{Unit correctness} \\
	Their generation approach should assign correct units to the variables (e.g. milk should be measured in liters, not in seconds)
	\item \textit{Coherence} \\
	The two variables introduced in their approach should be coherent to a joint context
\end{enumerate}

The system proposed by the authors takes as input only a user defined prompt. They choose to generate text conditioned on this prompt and apply constraints afterwards by identifying critical constituents (in this case words) via POS tagging and adjusting the tokens that are identified to invalidate constraints. This Checking strategy requires the manual definition of POS tag sequences as well as mathematical rules to identify relevant tokens and check for validity. The model that is used for text generation has been trained on a dataset of exercises similar to the desired output exercise. This limits their approach in that the structure of exercises generated by the approach is still very similar among exercises. In addition to this, the presented examples are missing punctuation and inflections are not always correct. A sample of exercises generated by their system is depicted in figure \ref{fig:generation-examples-rw}. 

\begin{figure}
	\begin{tightcenter}
		\begin{mdframed}
			\raggedright
			\textbf{Example 1} \\
			"Vimal built house and he used 21 kg cement and 20 l water, how much more cement than water did vimal use"
			
			\textbf{Example 2} \\
			"Supun wrote essay and he used 5 pens and 4 pencils, how much more pens than pencils did Supun use"
		\end{mdframed}	
	\end{tightcenter}
	\caption{Examples of exercises created by (\cite{liyanage2019multilanguage}, p. 5)} \label{fig:generation-examples-rw}
\end{figure}

In (\cite{liyanage2020multi}, p. 4709-4716) the same authors adapted their model to alleviate the need for manually defining the aforementioned checking strategy. They achieve this by incorporating POS tags into the training data. Thereby the updated model is able to learn the checking strategy itself, given sufficient training data. The exercises still remain rather similar in their constituent structure though, since the same training data has been used as for the previous model. 

The system proposed by the authors of (\cite{polozov2015personalized}, pp. 381-388) takes two sets of requirements as input. One set contains tutor requirements, e.g. the type of operations and equations an exercise shall contain. The second set contains student requirements, such as the context of the exercise, the actors and their relationships. 

Their approach first constructs a valid logical graph for an exercise and afterwards realizes this graph into text though a realization engine. In this context valid means that the generated logical graphs adhere to all requirements set by tutor and student. In addition to the user defined requirements the system aims to generate plots that tend to be interesting by introducing discourse tropes. Discourse tropes are facts that do not add to or interfere with the mathematical problem at hand but make the plot more engaging and lively.

This approach differs greatly from the previous approaches, since it is strongly logic based and uses no NLM components. It therefore requires far more details for generating an exercise in the form of structured user input as well as predefined ontologies. These ontologies are required for the model build its logical graphs and therefore pose a bottleneck in scaling the model to arbitrary contexts.

A similar, yet by far less complex approach incorporating OWL ontologies is proposed by (\cite{williams2011generating}, pp. 61-64).

Some authors employ template-based methods to generate questions. This approach entails with less diversity in their exercises (cf. e.g. \cite{deane2003automatic}, p. 6).

Finally, some consolidating observations are noted:
\begin{enumerate}
	\item All authors generate text exclusively via neural methods or templating methods
	\item Authors opting for neural text generation methods additionally pose the task of constraint satisfaction onto the NLM	
	\item All authors take as input exclusively either structured (e.g. equations) or unstructured (e.g. prompts) inputs. Those authors that only use unstructured input rely on their model to define structure based on the data it has seen during training time, therefore limiting the possible structures
\end{enumerate}

\newpage

\section{Proposed Method for Word Problem Generation}

\subsection{General Approach for generating Constrained Text}

\begin{figure}
	\centering
	\begin{tikzpicture}[x=0.5cm,y=0.5cm]
		\node at (10,7) (a) {};
		\node at (28,8) (b) {};
		\draw (8.97,6.5) |- ++(20.06,2) |- cycle;
		\node[rectangle,fit=(a) (b)] (x) {};
		\node[anchor=center,inner sep=3pt] at (x.center) {Generation controller};
		
		\node at (9.5,-2) (a) {};
		\node at (28.5,-3.5) (b) {};
		\node[draw=black,dashed,fill=black!10,rectangle,fit=(a) (b)] (x) {};
		
		\node at (10,-2.5) (a) {};
		\node at (18,-3) (b) {};
		\node[draw=black,fill=white,rectangle,fit=(a) (b)] (x) {};
		\node[anchor=center,inner sep=3pt] at (x.center) {Conflict controller};
		
		\node at (20,-2.5) (a) {};
		\node at (28,-3) (b) {};
		\node[draw=black,fill=white,rectangle,fit=(a) (b)] (x) {};
		\node[anchor=center,inner sep=3pt] at (x.center) {Coherence controller};
		
		\node at (9.5,-0.5) (a) {};
		\node at (28.5,5.5) (b) {};
		\node[draw=black,dashed,fill=black!10,rectangle,fit=(a) (b)] (x) {};
		
		\node at (10,0) (a) {};
		\node at (18,5) (b) {};
		\node[draw=black,fill=white,rectangle,fit=(a) (b)] (x) {};
		\node[anchor=north west,inner sep=3pt] at (x.north west) {Infill generator};
		\node at (x.south west) (s) {};
		\draw (s) ++(0.8,1.2) circle(1mm)[fill] ++(0.5,0) circle(1mm)[fill] ++(0.5,0) circle(1mm)[fill] ++(0.5,0) circle(1mm)[fill] ++(0.5,0) circle(1mm)[fill=black];
		\draw (s) ++(0.3,0.5) node[align=left, anchor=west] {\footnotesize Left};
		\node at (x.south east) (s) {};
		\draw (s) ++(-0.8,1.2) circle(1mm)[fill] ++(-0.5,0) circle(1mm)[fill] ++(-0.5,0) circle(1mm)[fill] ++(-0.5,0) circle(1mm)[fill] ++(-0.5,0) circle(1mm)[fill=black];
		\draw (s) ++(-3.3,0.5) node[align=left, anchor=west] {\footnotesize Right};
		\node at (x.south) (s) {};
		\draw (s) ++(-1,1.2) circle(1mm);
		\draw (s) ++(-0.5,1.2) circle(1mm);
		\draw (s) ++(0,1.2) circle(1mm);
		\draw (s) ++(0.5,1.2) circle(1mm);
		\draw (s) ++(1,1.2) circle(1mm);
		\draw (s) ++(-1.5,0.5) node[align=left, anchor=west] {\footnotesize Output};
		\node at (10.5,2.8) (a) {};
		\node at (17.5,3.8) (b) {};
		\node[draw=black,rectangle,fit=(a) (b),fill=myblue] (x) {};
		\node[anchor=center,inner sep=3pt] at (x.center) {Encoder-Decoder};
		
		\node at (x.south west) (s) {};
		\draw[color=black!15!white,line width=15] (s) ++(1.3,-0.4)-- ++(0,-0.8);
		\path[fill=black!15!white] (s) ++(1.3,-0.5) -- ++(1,0) -- ++ (-1,0.5) -- ++(-1,-0.5) -- ++(1,0);
		\node at (x.south) (s) {};
		\draw[color=black!15!white,line width=15] (s) ++(0,0)-- ++(0,-0.9);
		\path[fill=black!15!white] (s) ++(0,-0.8) -- ++(1,0) -- ++ (-1,-0.5) -- ++(-1,0.5) -- ++(1,0);
		\node at (x.south east) (s) {};
		\draw[color=black!15!white,line width=15] (s) ++(-1.3,-0.4)-- ++(0,-0.8);
		\path[fill=black!15!white] (s) ++(-1.3,-0.5) -- ++(1,0) -- ++ (-1,0.5) -- ++(-1,-0.5) -- ++(1,0);
		
		\node at (20,0) (a) {};
		\node at (28,5) (b) {};
		\node[draw=black,fill=white, rectangle,fit=(a) (b)] (x) {};
		\node[anchor=north west,inner sep=3pt] at (x.north west) {Rule-based generator};
		\node at (x.south west) (s) {};
		\draw (s) ++(0.8,1.2) circle(1mm)[fill] -- ++(0.6,0.8) circle(1mm)[fill] -- ++(0.3,-0.7) circle(1mm)[fill] ++(-0.3,0.7) -- ++(-0.55,0.1) circle(1mm)[fill] ++(+0.55,-0.1) -- ++(0.6,-0.25) circle(1mm)[fill] -- ++(-0.2,0.6) circle(1mm)[fill] -- ++(-0.4,-0.35);
		\draw (s) ++(0.3,0.5) node[align=left, anchor=west] {\footnotesize Struct};
		\node at (x.south east) (s) {};
		\draw (s) ++(-2.8,1.2) circle(1mm);
		\draw (s) ++(-2.3,1.2) circle(1mm);
		\draw (s) ++(-1.8,1.2) circle(1mm);
		\draw (s) ++(-1.3,1.2) circle(1mm);
		\draw (s) ++(-0.8,1.2) circle(1mm);	
		\draw (s) ++(-3.3,0.5) node[align=left, anchor=west] {\footnotesize Output};
		\node at (20.5,2.8) (a) {};
		\node at (27.5,3.8) (b) {};
		\node[draw=black,rectangle,fit=(a) (b),fill=mygreen] (x) {};
		\node[anchor=center,inner sep=3pt] at (x.center) {Decoder};
		
		\node at (x.south east) (s) {};
		\draw[color=black!15!white,line width=15] (s) ++(-1.3,0)-- ++(0,-0.9);
		\path[fill=black!15!white] (s) ++(-1.3,-0.8) -- ++(1,0) -- ++ (-1,-0.5) -- ++(-1,0.5) -- ++(1,0);
		\node at (x.south) (s) {};
		\draw[color=black!15!white,line width=10] (s) ++(-1,-0.4)-- ++(0,-0.8) -- ++(-1,0);
		\path[fill=black!15!white] (s) ++(-1,-0.5) -- ++(0.6,0) -- ++ (-0.6,0.5) -- ++(-0.6,-0.5) -- ++(0.6,0);
	\end{tikzpicture}
	\caption{Model architecture and components (own illustration)} \label{fig:architecture-approach}
\end{figure}
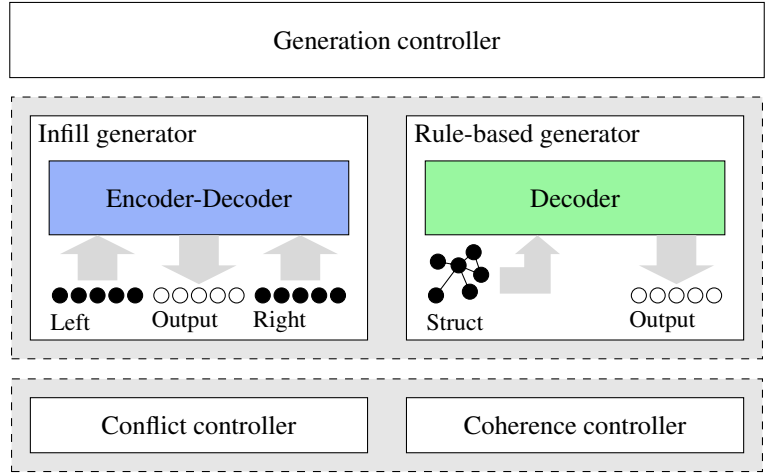

The key design choice of the presented approach is to disentangle the task of generating validity constrained text into generating distinct text constituents and combining them in a controlled manner. This allows for the introduction of a generation controller that is able to adapt the generation strategy depending on whether constituents must adhere to content constraints or not. Specifically, this distinction is defined to take place on sentence-level. The overall architecture is presented in figure \ref{fig:architecture-approach}.

The proposed model takes as input two types of user provided data. Unstructured user input, namely a prompt for setting the context of the exercise to be generated, and structured user input, encoding the exercise specification in terms of exercise type and hardness.

Constrained generation can rely on the determinateness of template-based approaches, i.e. realization engines, guaranteeing constraint satisfaction at the cost of restricting diversity.  Though, through designing templates in an adequate manner, diversity is reasonably retained. This type of generation is based completely on structured user input. Unconstrained generation on the other hand can harness the creativity of Neural Language Generation (NLG) approaches, i.e. NLMs, to produce diverse content. This type of generation is mostly based on unstructured user input.

Combining both constrained and unconstrained constituents is facilitated through first arranging them in a varying order via the generation controller and generating connecting text segments to merge constituents. The constituents being generated in this manner provide semantic continuity to the overall text. The arrangement of constituents as well as the amount of additional context incorporated into an exercise is dependent on the desired hardness of the exercise.

\begin{figure}[t]
	\centering
	\begin{tikzpicture}[x=0.5cm,y=0.5cm]
		\node at (0,0) (a) {};
		\node at (28,8) (b) {};
		\node[draw=black,rectangle,fit=(a) (b)] (x) {};
		\node[anchor=north west,inner sep=3pt] at (x.north west) {Exercise text};
		\path (x.north west) ++(0.4,-1) node[anchor=north west,inner sep=3pt, text width=396, align=justify] {\baselineskip=5pt Berlin is the German city with the most traffic lights per capita. \hlblue{This fact already tells you that there will be lots of traffic jams during the rush hour.} \hlblue{Regardless of this fact traffic lights behave like ordinary ones.} \hlgreen{A traffic light is red with a chance of 55\%.} \hlgreen{Furthermore, a traffic light is either red or green.} \hlblue{Additionally the following holds:} \hlgreen{Every traffic lights chances of being either red or green are independent.} \par
			
			\hlgreen{\linebreak	Calculate the chance that 5 out of 8 traffic lights are red.}};

	\end{tikzpicture}
	\caption{Example exercise highlighting the origin of constituents (own illustration)} \label{fig:exercise-example}
\end{figure}
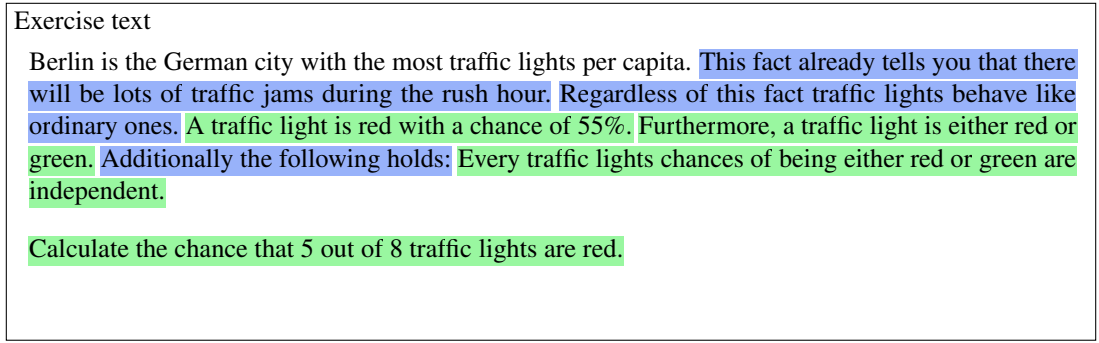

To clarify the previously introduced concepts, figure \ref{fig:exercise-example} displays an example exercise including labeled constituents. Color indicates which generator was used, no color means user input is adopted without change.

Constituents generated via NLMs are programmatically checked for constraint satisfaction by a conflict controller. Concretely, every non-constrained constituent is checked against every constrained constituent via Natural Language Inference (NLI) methods. Since constrained constituents are guaranteed to satisfy all constraints, they can be used as a ground truth assumption. If a contradiction is detected, the generation controller discards and replaces the constituent. This way all constituents are guaranteed to be valid. 

Contrasting this approach to the approaches presented in the related work section, several new ideas are prevalent:
\begin{enumerate}
	\item The task of text generation is split into template based approaches and NLG	
	\item Constraint satisfaction is not directly incorporated into the NLM but rather outsourced to a specialized NLI model	
	\item The user can directly influence the generation style of the system instead of giving full control to a pre-trained NLM, allowing for e.g. adjustable hardness of exercises 
\end{enumerate}

The pursuit of modularity has been a top priority in designing the proposed approach. One positive aspect of this modularity is the fact that individual building blocks can be scaled, improved and exchanged flexibly. Thereby, one can freely interchange implementations of each building block  to keep up with the State of the Art (SOTA) of the respective subtasks. Doing the same with an end-to-end model would require a full retraining.

Finally, having generated constrained constituents in a determinate manner allows for the simultaneous generation of a structured representation of the exercise in addition to its textual form. This in turn enables downstream tasks such as solution explanation and error highlighting to maximize learning effectivity. Again, such a structured representation is not easily extractable when utilizing a single end-to-end NLM.

\subsection{Technical Details}

The following sections outline the technical details of the implemented Proof of Concept (POC). Each section focuses on an individual module of the proposed method and highlights key design choices. 
The POC supports the generation of exercises that cover both the Binomial distribution as well as the Normal distribution. Both exercise types will serve as examples in the upcoming section.\footnote{An interactive demo of the system is provided online at \href{https://colab.research.google.com/drive/1TXZViybIiwRhj_7usPyfWWOIuRmsxnza?usp=sharing}{Google Colab}}

\subsection{Data Input}

The data input to the POC is implemented as a simple command line interface (CLI). As outlined above two types of input are provided by the user, an exercise context and an exercise specification.
The exercise context is provided through a user prompt that sets the scene for the exercise. This prompt is the basis for generating constraint constituents and embedding them in a joint context. 

Based on the initial context the user provides an exercise specification containing all required information for constrained generation. This is facilitated through a guided dialog that interactively requests necessary information from the user depending on the exercise type. The inputs to this dialog are process via SimpleNLG, yielding non-canned constituents. These constituents allow for correct inflections when realizing text.

Finally, the user instructs the generation controller on how to generate the exercise. To preserve a lean interface to the user only one parameter needs to be set in this regard, namely the exercise hardness. This parameter in turn affects multiple hyperparameters of the generation controller i.e., question sampling, constituent generation and constituent arrangement.

\begin{figure}[t]
	\centering
	\includegraphics[scale=0.27]{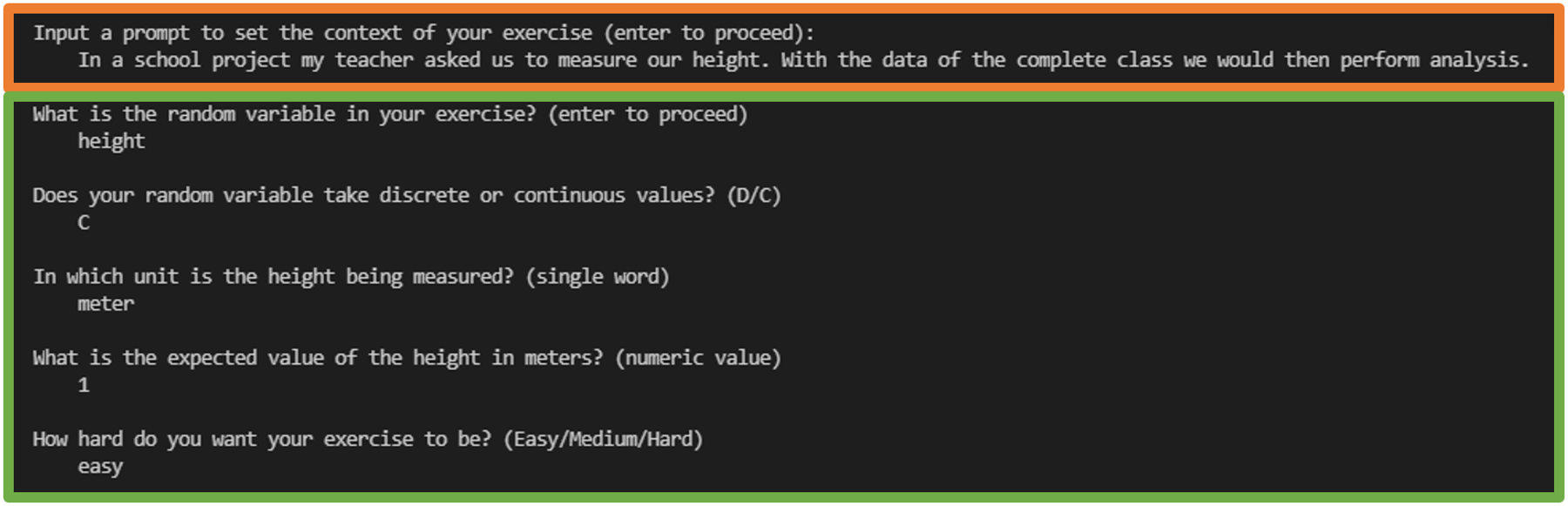}
	\caption{Example CLI input dialog (own illustration)} \label{fig:cli-example}
\end{figure}

An exemplary input dialog is displayed in figure \ref{fig:cli-example}. Unstructured input is highlighted in orange color, structured input in green color.

\newpage

\subsection{Generation Controller}

The generation controller takes as input the prompt and exercise specification from the input dialog. The general generation strategy is presented in consists of the following steps:
\begin{enumerate}
	\item Generate constraint constituents	
	\item Arrange constraint constituents and placeholders for connecting constituents	
	\item Generate connecting constituents and insert into placeholders
	\item Check constituent validity
	\item Check constituent coherence
\end{enumerate}

Based on the amount of constraint constituents being generated as well as their precedence constraints the constituents are arranged in a random fashion. Concretely, the employed algorithm for sampling arrangements is presented in figure \ref{fig:arrangement-function}.

The arrangement is managed as a list in which each item is associated with a constituent. For example, the item 'P' is linked to the user specified prompt and the item 'I' is a placeholder for a connecting constituent. Every variable prefixed by 'self' are hyperparameters of the generation controller that are indirectly controlled by the user through the hardness parameter. Lines 2-4 randomize the amount of connecting constituents between the user prompt and the first constrained constituent. Python3's native 'random' package is used for randomization. Lines 6-19 arrange the remaining constrained constituents and insert placeholders for connecting constituents in a random fashion. Lines 21-25 exchange the items associated with statements with statement indices. These are used by the generation controller to randomize the statement order.

As previously hinted to, hardness parameter affects both the sampling of questions and the arrangement of constituents. 
The arrangement is affected by appropriately setting the parameters outlined in the sampling algorithm outlined above. A hard exercise tends to contain more connecting constituents, with relevant information being scattered sparsely among them. In contrast, easy exercises tend to present relevant information to the student without a lot of obfuscation. Additionally, a hard exercise samples more and harder questions from the questionpool than an easy exercise.

The generation of the respective constituents as well as the content checks performed by the generation controller will be outlined in the following sections.

\subsection{Constrained Generation}

Since the POC focuses on the use-case of generating word problems for mathematical statistics, two constituent types are identified to require constraint generation:

\begin{enumerate}
	\item \textit{Statements} \\
	Statements include all information that is required for the student to successfully solve the exercise. In the context of mathematical statistics, this includes all assumptions needed to identify both the relevant random variable as well as its underlying statistical distribution. Additionally, all relevant distribution parameters need to be included, either implicitly or explicitly.
	\item \textit{Questions} \\
	Questions include all questions related to an exercise context. These questions need to be consistent with the context so as to make sure that they can be answered without any additional information.
\end{enumerate}

\begin{figure}
\begin{python}
	def get_arrangement(self, statements):
	arrangement = []
	arrangement += 'P'
	arrangement += 'I'*r.choice(range(1,self.max_prefix_ext_sents+1))
	
	rem_chunks = self.max_statement_chunks
	rem_statements = len(statements)
	while rem_chunks>self.min_statement_chunks and rem_statements>0:
	amt = random.choice(range(1,rem_statements+1))
	arrangement += 'S'*amt
	arrangement += 'I'*r.choice(range(1,self.max_infill_sents+1))
	rem_chunks -= 1
	rem_statements -= amt
	while rem_statements > 0:
	arrangement += 'S'
	arrangement += 'I'*r.choice(range(1,self.max_infill_sents+1))
	rem_statements -= 1
	while arrangement[-1] == 'I':
	arrangement = arrangement[:-1]
	
	arrangement = list(arrangement)
	i = 0
	while 'S' in arrangement:
	arrangement[arrangement.index('S')] = str(i)
	i += 1
	
	return arrangement\end{python}
\caption{Implementation of the constituent arrangement function in the POC (own code)} \label{fig:arrangement-function}
\end{figure}

The distinction between statements and questions is made because they are affected differently by the constraints of the use-case (Table \ref{tab:requirements-examples} on page \pageref{tab:requirements-examples}). Statements need to be embedded in the exercise context, whereas questions are disentangled from the context.

Enforcing constraints on statements is required to guarantee completeness and correctness. Additionally, they indirectly allow for enforcing consistency and unequivocality by providing a ground truth to check all remaining constituents against.

Questions, in contrast, are subject only to satisfying correctness and unequivocality. Completeness does not apply since there is an infinite set of possible questions and consistency is not required since each question is defined to be independent of others.

As a part of the guided input dialog the user implicitly specifies the type of exercise to be generated. As the POC only supports two exercise types, this specification solely depends on whether the random variable takes on discrete or continuous values. Depending on this input, specific statements and questions for each exercise type are generated via the SimpleNLG generation engine.

An exercise class is instantiated with a predefined set of canned and non-canned constituents. These are then integrated into predefined randomized templates of both statements and questions. These templates are then realized to human readable text. The amount of statements and their semantic content is always identical across exercise instances of the same type since the information required to an exercise type is always the same. To introduce additional variance besides randomized templates, the order of statements is varied.

Questions on the other hand can differ substantially depending on the hardness of an exercise. Therefore, instead of a fixed set of questions, a questionpool is generated. This questionpool consists of ordered sets $(Q,H)$ where $Q$ is a realized question and $H$ is its assigned hardness level. This allows the generation controller to sample questions from this questionpool depending on the desired exercise hardness. The order of the questions is varied under the precedence constraint that easier questions should precede harder questions.

\subsection{Unconstrained Generation}

All connecting constituents, also referred to as connections, are generated in an unconstrained fashion via NLG methods. To ensure fluency in the generated exercise, these constituents need to be coherent with their preceding and succeeding constituents. To incorporate both left and right context when generating a constituent standard NLM approaches need to be augmented. Among various approaches presented in the literature (cf. e.g. (\cite{ijcai2019-727}, pp. 5233-5239), (\cite{mallinson2020felix}, pp. 2-4)) the author opts for an approach based on (\cite{donahue2020enabling}, pp. 2-3) since it empirically produces the most viable results in terms of content coherence.

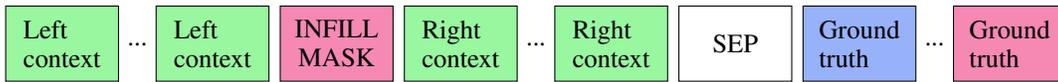
\begin{figure}
	\begin{tightcenter}
		\begin{tikzpicture}
			
			\node[anchor=west] at (-4.5,-3) {\textbf{Infill language modelling}};
			
			\draw[fill=mygreen]  (-4.5,-3.5) rectangle (-3,-4.5);
			\node[align=left] at (-3.75,-4) {Left\\context};
			
			\node at (-2.75,-4) {...};
			
			\draw[fill=mygreen]  (-2.5,-3.5) rectangle (-1,-4.5);
			\node[align=left] at (-1.75,-4) {Left\\context};
			
			\draw[fill=myred]  (-0.85,-3.5) rectangle (0.65,-4.5);
			\node[align=center] at (-0.1,-4) {INFILL\\MASK};
			
			\draw[fill=mygreen]  (0.8,-3.5) rectangle (2.3,-4.5);
			\node[align=left] at (1.55,-4) {Right\\context};
			
			\node at (2.55,-4) {...};
			
			\draw[fill=mygreen]  (2.8,-3.5) rectangle (4.3,-4.5);
			\node[align=left] at (3.55,-4) {Right\\context};
			
			\draw  (4.45,-3.5) rectangle (5.95,-4.5);
			\node at (5.2,-4) {SEP};
			
			\draw[fill=myblue]  (6.1,-3.5) rectangle (7.6,-4.5);
			\node[align=left] at (6.85,-4) {Ground\\truth};
			
			\node at (7.85,-4) {...};
			
			\draw[fill=myred]  (8.1,-3.5) rectangle (9.6,-4.5);
			\node[align=left] at (8.85,-4) {Ground\\truth};
			
		\end{tikzpicture}
	\end{tightcenter}
	\caption{ILM training objective (own illustration)}\label{fig:ilm-training-record}
\end{figure}

Donahue et al. finetune an instance of GPT-2 to predict masked spans with varying lengths in an input context. Specifically, they use spans masking paragraphs, sentences, ngrams and single words. They call this training objective Infill Language Modelling (ILM). An exemplary record of their training dataset is shown in figure \ref{fig:ilm-training-record}. The current token to be predicted is highlighted in blue color, context tokens that can be considered by the model are highlighted in green color and masked tokens are highlighted in red color.

By casting the problem of text infilling into this structure, a classical NLM is able to incorporate the context on both sides of the masked segment and its position in the context into its autoregressive generation of tokens. That way, the creative NLG capabilities of pre-trained NLMs like GPT-2 can be harnessed in generating connecting segments of text.

For the POC, their model is specifically finetuned on only considering sentences with left and right context. Additionally, the special mask tokens for predicting paragraphs, ngrams and words are omitted so that the model solely learns to generate a single sentence. This allows the generation controller to give precise instructions on how many connecting constituents shall be generated. The custom finetuning dataset is based on the ROCStories dataset (cf. \cite{mostafazadeh2016corpus}, pp. 3-6).

\subsection{Constraint Checking}

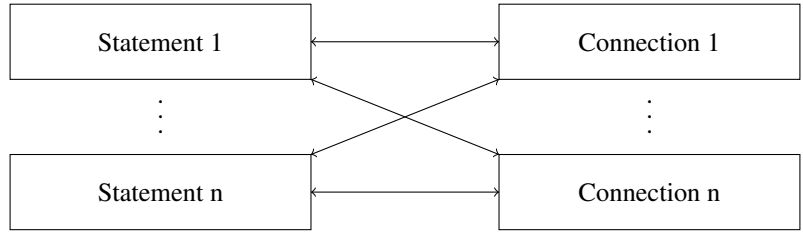
\begin{figure}
	\begin{tightcenter}
		\begin{tikzpicture}
			
			\draw  (-5.5,2) rectangle (-1.5,1);
			\node at (-3.5,1.5) {Statement 1};		
			
			\draw  (-5.5,0) rectangle (-1.5,-1);
			\node at (-3.5,-0.5) {Statement n};		
			
			\draw  (1,2) rectangle (5,1);
			\node at (3,1.5) {Connection 1};		
			
			\draw  (1,0) rectangle (5,-1);
			\node at (3,-0.5) {Connection n};			
			
			\draw[<->] (-1.5,-0.5) -- (1,-0.5);
			\draw[<->] (-1.5,1.5) -- (1,1.5);
			\draw[<->] (-1.5,-0) -- (1,1);
			\draw[<->] (-1.5,1) -- (1,-0);
			\node[rotate=90] at (-3.5,0.5) {. . .};
			\node[rotate=90] at (3,0.5) {. . .};
		\end{tikzpicture}
	\end{tightcenter}
	\caption{Method of constraint checking in the POC (own illustration)} \label{fig:constraint-checking-approach}
\end{figure}

Constraint checking is facilitated via an NLI model that checks for conflicts between those constituents that have been generated in an unconstrained fashion and those that have been generated in a constrained fashion. Specifically, every inter-constituent relationship is checked, as shown in figure \ref{fig:constraint-checking-approach}. This method is chosen since, even though one would attribute transitivity properties to conflicts between sentences, current NLI methods are not robust enough to reliably detect all conflicts when only checking a reduced inter-constituent graph.

For the implementation of the POC the author chooses a DeBERTa (\cite{he2021deberta}, pp. 3-5) encoder for the sole reason that it achieves SOTA performance in numerous NLI tasks. As outlined in the introduction of this chapter, modularity allows for easy integration of a different model when the SOTA advances. The algorithm for constraint checking is presented in figure \ref{fig:constraint-checking}.

\begin{figure}
	\begin{python}
		def check_constraint_conflicts(statements, candidates):
			conflict = False
			for i,c in enumerate(candidates):
				if conflict:
					break
				for s in statements:
					x = nli_tokenizer.encode(s, c, return_tensors='pt')
					probs = nli_model(x)[0][0].softmax(dim=0)
					
					entailment = probs[2].item()
					neutral = probs[1].item()
					contradiction = probs[0].item()
					
					conflict = contradiction > .5 and entailment < .2
					if conflict:
						break
				
			return not conflict\end{python}
	\caption{Implementation of constraint checking function in the POC (own code)} \label{fig:constraint-checking}
\end{figure}

The constraint checking function expects both a ground truth, namely the statements orginitating from constrained generation, and candidates to be checked against this ground truth, namely the connections originating from unconstrained generation, as input. It checks each edge of the inter-constituent graph via two loops and detects conflicts based on a predefined condition. Since the employed NLI model outputs probabilities for each of the three possible classes 'entailment', 'neutral' and 'conflict', a strategy on detecting true conflicts based on these probabilities needs to be defined. The POC employs the condition defined on line 14 of figure \ref{fig:constraint-checking}. The optimal parameters of this condition have been identified in the course of conducting the experiments outlined in section 4. The function returns a boolean value that indicates whether any conflict has been detected. Additionally, early stopping is implemented for improved performance. That way, if a conflict is detected on checking the first edge of the conflict graph, no further edges are checked and the function returns its output early.

\begin{figure}
	\begin{tightcenter}
		\begin{tikzpicture}
			\node at (0,0) (a) {};
			\node at (14,3.5) (b) {};
			\node[draw=black,rectangle,fit=(a) (b)] (x) {};
			\node[anchor=north west,inner sep=3pt] at (x.north west) {Exercise text};
			\path (x.north west) ++(0.2,-0.5) node[anchor=north west,inner sep=3pt, text width=396, align=justify] {\baselineskip=5pt I am currently very busy writing my thesis. I am writing many pages every day. But sometimes I think I am not as productive as I could be. \hlblue{Or I feel I am not as productive as I once was.} \hlgreen{Days are productive with a chance of 75\%.} \hlgreen{A day only is either lazy or productive.} \hlblue{We argue a lot that the chance is closer to 90\%.} \hlgreen{Every days probabilities of being either lazy or productive are independent.}};
			
			\draw[fill=mygreen]  (-0.25,-1.05) rectangle (6,-2.05);
			\node[align=left, anchor=west] at (-0.25,-1.55) {Days are productive with a\\ chance of 75\%.};		
			
			\draw[fill=mygreen]  (-0.25,-2.55) rectangle (6,-3.55);
			\node[align=left, anchor=west] at (-0.25,-3.05) {A day only is either lazy or\\ productive.};	
			
			\draw[fill=mygreen]  (-0.25,-4.05) rectangle (6,-5.05);
			\node[align=left, anchor=west] at (-0.25,-4.55) {Every days probabilities of being\\ lazy or productive are independent.};
			
			\draw[fill=myblue]  (8,-1.55) rectangle (14.25,-2.55);
			\node[align=left, anchor=west] at (8,-2.05) {Or I feel I am not as productive as\\ I once was.};		
			
			\draw[fill=myblue]  (8,-3.55) rectangle (14.25,-4.55);
			\node[align=left, anchor=west] at (8,-4.05) {We argue a lot that the chance\\ is closer to 90\%.};			
			
			\draw[<->] (6,-3.55) -- (8,-4.05);
			\draw[<->] (6,-1.55) -- (8,-1.55);
			\draw[<->] (6,-2.55) -- (8,-2.05);
			\draw[<->] (6,-4.55) -- (8,-4.55);
			\draw[<->] (6,-4.05) -- (8,-2.55);
			\draw[<->, draw=red,line width=0.5mm] (6,-2.05) -- (8,-3.55);
		\end{tikzpicture}
	\end{tightcenter}
	\caption{Exemplary conflict check performed on a sample exercise (own illustration)} \label{fig:constraint-checking-example}
\end{figure}
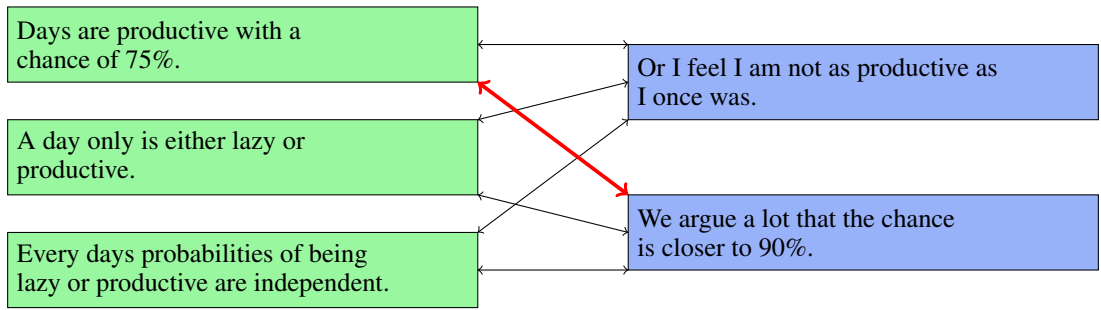

An exemplary conflict check is presented in figure \ref{fig:constraint-checking-example}. Conflicting edges are highlighted in red color.

\subsection{Coherence Checking} \label{Coherence Checking}

Coherence checking is performed without a distinction between the generation strategy of individual constituents since it focuses solely on semantic fluency. Specifically, only local coherence is checked, meaning that only adjacent constituents need to be coherent. 

\begin{figure}
	\begin{tightcenter}
		\begin{tikzpicture}
			
			\draw  (-5.5,2) rectangle (-1.5,1);
			\node at (-3.5,1.5) {Constituent n-2};		
			
			\draw  (-5.5,0.5) rectangle (-1.5,-0.5);
			\node at (-3.5,0) {Constituent n-1};		
			
			\draw  (-5.5,-1) rectangle (-1.5,-2);
			\node at (-3.5,-1.5) {Constituent n};		
			
			\draw[->]  plot[smooth, tension=.9] coordinates {(-1.5,1.25) (-0.75,0.75) (-1.5,0.25)};
			\draw[->]  plot[smooth, tension=.9] coordinates {(-1.5,-0.25) (-0.75,-0.75) (-1.5,-1.25)};
		\end{tikzpicture}
	\end{tightcenter}
	\caption{Method of coherence checking in the POC (own illustration)} \label{fig:coherence-checking-approach}
\end{figure}
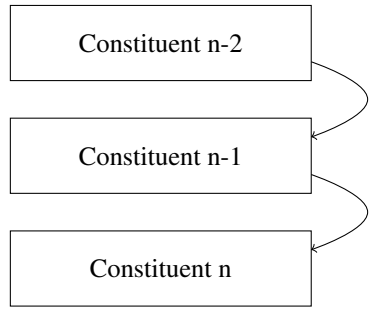

This decision is grounded in the assumption that by providing local coherence sufficient fluency is provided to an exercise. Additionally, since constituents generated in a constraint fashion most definitely are coherent with the prompt of the exercise and one another, enforcing local coherence around those scattered constituents implicitly leads to some degree of global coherence. The coherence checking procedure is displayed in figure \ref{fig:coherence-checking-approach}.

The POC combines two measures of coherence, semantic similarity and sentence order. Semantic similarity is measured in the cosine distance of sentence embeddings. These embeddings are calculated by a finetuned MPNet encoder (\cite{song2020mpnet}, pp. 2-6), chosen for its SOTA performance. For assessing sentence order a standard BERT encoder (\cite{devlin2019bert}, pp. 3-5) is finetuned on ROCStories (\cite{mostafazadeh2016corpus}, pp. 3-6) with only next sentence prediction (NSP) as the training objective. This finetuned model is then used to assess the sentence order between constituents. The concrete implementation of the coherence checking function in the POC is presented in figure  \ref{fig:coherence-checking-function}. 

\begin{figure}
	\begin{python}
		def check_consecutive_coherence(sents, arrangement):
			coherence = True
			for i in range(len(sents)-1):
				s1 = sents[i]
				s2 = sents[i+1]
				encoding = coh_tokenizer(s1, s2, return_tensors='pt')
				outputs = coh_model(**encoding, labels=torch.LongTensor([1]))
				logits = outputs.logits.softmax(dim=1)
				if logits[0, 0] < 0.99:
					coherence = False
					break
				encoding1 = cos_model.encode(s1, convert_to_tensor=True)
				encoding2 = cos_model.encode(s2, convert_to_tensor=True)
				cos_dist = util.pytorch_cos_sim(encoding1, encoding2)
				if cos_dist.item() > 0.3:
					coherence = False
					break
			return coherence\end{python}
	\caption{Implementation of coherence checking function in the POC (shortened, own code)} \label{fig:coherence-checking-function}
\end{figure}

As outlined in figure \ref{fig:coherence-checking-approach} coherence is checked via a single forward pass through all constituents. This is implemented in a single loop starting in line 3. In each iteration consecutive constituents are compared in terms of sentence ordering (lines 6-11) and semantic similarity (lines 12-17). The sentence ordering task requires a joint encoding of both constituents, whereas the semantic similarity task encodes both constituents independently. This difference is merely an architectural choice of the used models. Again, the incoherence detection conditions for both metrics are determined to be optimal through experiments in section 4.

\begin{figure}
	\begin{tightcenter}
		\begin{tikzpicture}
			\node at (0,0) (a) {};
			\node at (14,4) (b) {};
			\node[draw=black,rectangle,fit=(a) (b)] (x) {};
			\node[anchor=north west,inner sep=3pt] at (x.north west) {Exercise text};
			\path (x.north west) ++(0.2,-0.5) node[anchor=north west,inner sep=3pt, text width=396, align=justify] {\baselineskip=5pt The average human drinks too little water per day. This phenomenon in conjunction with an unhealthy diet has led to high obesity in western countries. \hlblue{Research shows that diets rely on water for cleansing the human body.} \hlgreen{Experience has shown that water consumptions can be assumed to follow a normal distribution.} \hlblue{People usually add occasion to massages to show the Eskimo leanness.} \hlgreen{The standard deviation in water consumption is assumed to be approximately 24.0 ml.} \hlgreen{The expected value water consumption is generally known to be $\sim$428 ml.}};
			
			\draw[fill=myblue]  (-0.25,-0.7) rectangle (13,-1.7);
			\node[align=left, anchor=west] at (-0.25,-1.2) {Research shows that diets rely on water for cleansing the human body.};	
			
			\draw[fill=mygreen]  (-0.25,-1.95) rectangle (13,-2.95);
			\node[align=left, anchor=west] at (-0.25,-2.5) {Experience has shown that water consumptions can be assumed to follow\\ a normal distribution.};		
			
			\draw[fill=myblue]  (-0.25,-3.2) rectangle (13,-4.2);
			\node[align=left, anchor=west] at (-0.25,-3.7) {People usually add occasion to massages to show the Eskimo leanness.};	
			
			\draw[fill=mygreen]  (-0.25,-4.45) rectangle (13,-5.45);
			\node[align=left, anchor=west] at (-0.25,-4.95) {The standard deviation in water consumption is assumed to be\\ approximately 24.0 ml.};	
			
			\draw[fill=mygreen]  (-0.25,-5.7) rectangle (13,-6.7);
			\node[align=left, anchor=west] at (-0.25,-6.2) {The expected value water consumption is generally known to be $\sim$428 ml.};
			
			\draw[->]  plot[smooth, tension=.9] coordinates {(13,-1.4) (13.75,-1.75) (13,-2.15)};
			\draw[->, draw=red,line width=0.5mm]  plot[smooth, tension=.9] coordinates {(13,-2.65) (13.75,-3) (13,-3.4)};
			\draw[->, draw=red,line width=0.5mm]  plot[smooth, tension=.9] coordinates {(13,-3.9) (13.75,-4.25) (13,-4.65)};
			\draw[->]  plot[smooth, tension=.9] coordinates {(13,-5.15) (13.75,-5.5) (13,-5.9)};
			
		\end{tikzpicture}
	\end{tightcenter}
	\caption{Exemplary coherence check performed on a sample exercise (own illustration)} \label{fig:coherence-checking-example}
\end{figure}
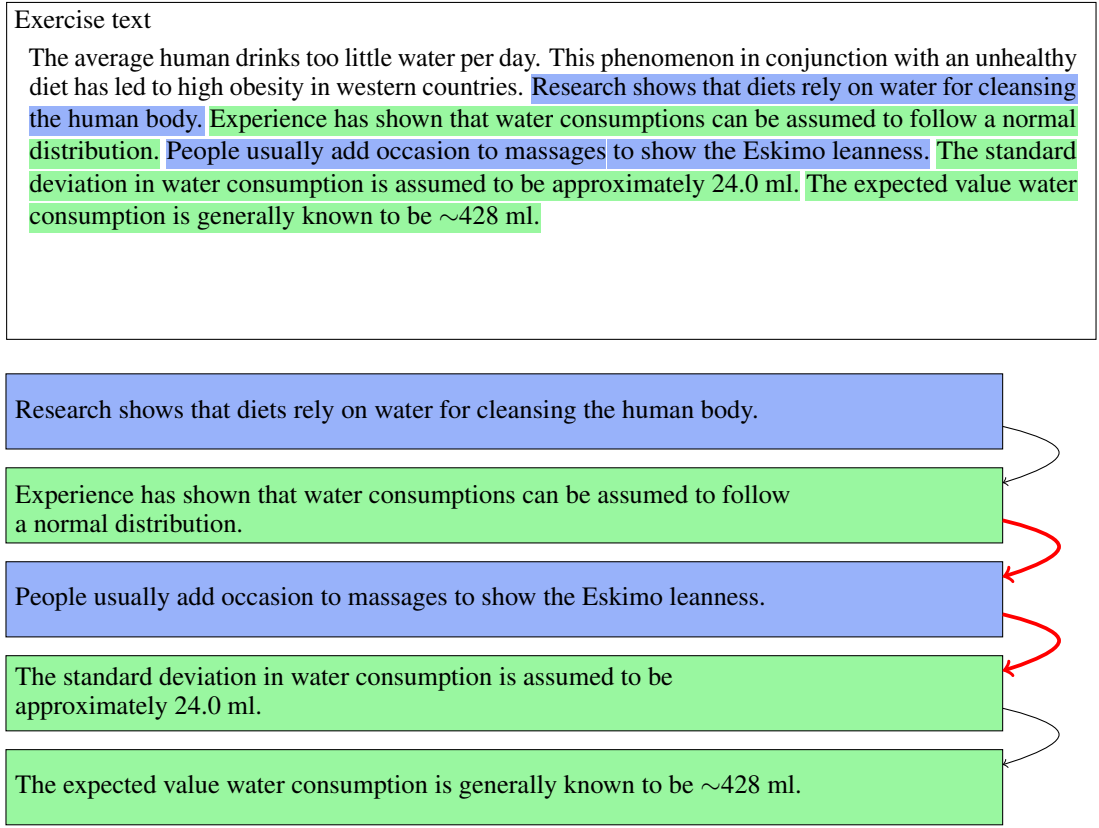

An exemplary coherence check is presented in figure \ref{fig:coherence-checking-example}. Incoherent edges are highlighted in red color.

\section{Experiments}

\subsection{Experimental Setup}

Recalling the research objectives defined in section \ref{Contributions of this thesis}, the effectiveness of the proposed method shall be testified in three dimensions.

\textit{Dimension 1 - Quantification of the validity of generated exercises}\\
To testify the validity of the exercises being generated a set of 100 exercises is being generated and checked for validity by humans. Specifically, the previously defined criteria for the validity of word problems in mathematical statistics apply.

Dimension 1 is captured via the metric $valid_\% $, namely the fraction of exercises that have been manually rated as valid by humans considering the set of 100 exercises.

\textit{Dimension 2 - Quantification of the coherence of generated exercises}\\
In a similar fashion the same 100 exercises are rated for coherence. In this context coherence specifically refers to the concept of semantic fluency introduced in section \ref{Coherence Checking}.

Dimension 2 is captured via the metric $coherent_\% $, defined analogously to $valid_\% $.

\textit{Dimension 3 - Quantification of the diversity of generated exercises}\\
For the purpose of quantification, diversity is defined as the heterogeneity in constituent arrangements. If the model can generate heterogenous arrangements and does not fall back to simple arrangements for every exercise, it is considered to be able to generate diverse exercises.
Content generation diversity is not being rated since it is mostly predefined by the exercise prompt and definition and thereby up to the user. What is rated though is the methods capability to handle diverse contents provided by the user. This is done to prevent the model from being great at generating exercises only if their content is about foxes, but fail when their content is about cars.

Both aspects of dimension 3 are captured in terms of entropy, defined as
\begin{align*}
	H = -\sum_{i=1}^{N} p(x_i) * log(p(x_i))
\end{align*}
with $[x_1,\cdots,x_N]$ being individual realizations of a random variable, $p(x_i)$ being the realization probability and $log$ being a standard base 2 logarithm. In the use-case at hand, the random variable might be the arrangement of constituents that can take different forms. Entropy is, put in layman's terms, a measure of expected surprise when sampling a random variable. The higher the expected surprise, the higher the entropy. It therefore serves as a viable metric for heterogenity in realizations.

Consequently, dimension 3 is captured via the metrics $arrangement_H$ and $content_H$.

In addition to the responses for each dimension, a proxy for the time taken to generate the representative set of 100 exercise instances is measured. $success_\%$ is defined as the percentage of generated exercises that the system labels as valid. For example, if $success_\% = 10\%$ holds, then for each supposedly valid exercise 10 exercises need to be generated, implying computation overhead and waiting times.

The model configuration is varied in several aspects to find the optimal parameters for diverse exercise generation. Namely, the following parameters are varied:
\begin{enumerate}
	\item \textit{$generation_{hardness}$} \\
	The exercise hardness affects both the hardness of questions and the complexity of the constituent structure
	\item \textit{$generation_{nucleus}$} \\
	The nucleus parameter of the decoding strategy for unconstrained generation
	\item \textit{$conflict_{condition}$} \\
	The condition for detecting conflicts from the three class probabilities of the NLI model (Figure \ref{fig:constraint-checking}, line 14)
	\item \textit{$coherence_{NSP}$} \\
	The condition for detecting incoherencies from NSP score (Figure \ref{fig:coherence-checking-function}, line 9)
	\item \textit{$coherence_{cosdist}$} \\
	The condition for detecting incoherencies from cosine distance score (Figure \ref{fig:coherence-checking-function}, line 15)
\end{enumerate}

The dataset created for testing contains records consisting of a prompt as well as its respective exercise configuration. The contents of these prompts have been chosen arbitrarily with heterogeneity in mind.

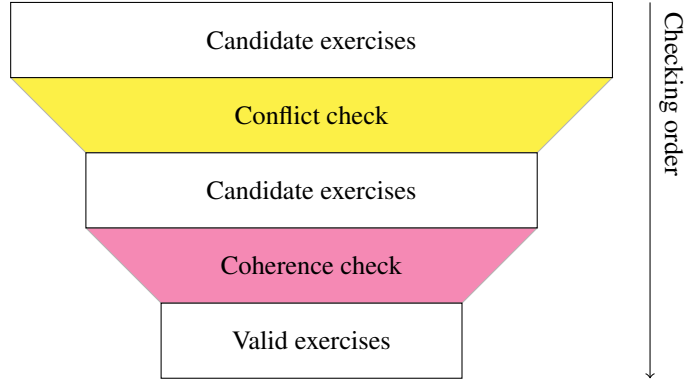
\begin{figure}
	\begin{tightcenter}
		\begin{tikzpicture}	
			\draw[fill=myyellow, draw=black!30] (-7.5,1) -- (0.5,1) -- (-0.5,0) -- (-6.5,0) -- (-7.5,1);
			
			\draw[fill=myred, draw=black!30] (-6.5,-1) -- (-0.5,-1) -- (-1.5,-2) -- (-5.5,-2) -- (-6.5,-1);
			
			\draw  (-7.5,2) rectangle (0.5,1);
			\node at (-3.5,1.5) {Candidate exercises};	
			
			\node at (-3.5,0.5) {Conflict check};	
			
			\draw  (-6.5,0) rectangle (-0.5,-1);
			\node at (-3.5,-0.5) {Candidate exercises};		
			
			\node at (-3.5,-1.5) {Coherence check};	
			
			\draw  (-5.5,-2) rectangle (-1.5,-3);
			\node at (-3.5,-2.5) {Valid exercises};		
			
			\draw[->] (1,2) -- (1,-3);
			\node[anchor=south west, rotate=270] at (1,2) {Checking order};
		\end{tikzpicture}
	\end{tightcenter}
	\caption{Exercise checking funnel considered for ablation study (own illustration)} \label{fig:funnel}
\end{figure}

In order to better understand the contributions of individual modules an ablation study is performed. In doing so, individual modules in the exercise checking funnel are removed temporarily. The two relevant modules are highlighted in figure \ref{fig:funnel}. 

All aforementioned tools and datasets are available at \url{https://github.com/Stansuro/ExGen}.

\subsection{Results}

Table \ref{tab:corr} presents the measured correlation coefficients of a $(parameter, response)$ pair. Every coefficient that satisfies $|r|>0.30$ is regarded as significant and a derivation is stated. The connection between measurement and derivation is defined in table \ref{tab:obs}. The following section outlines the aforementioned derivations.

\begin{enumerate}
	\item \textit{More diverse generation entails more waiting time} \\
	A higher value for the $generation_{nucleus}$ instructs the NLM to generate more creative text. This leads to more exercises that are labelled as invalid by the system, therefore entailing lower $success_\%$
	\item \textit{Less diverse generation entails more coherent constituents} \\
	A lower value for the $generation_{nucleus}$ instructs the NLM to generate less creative text. This leads to less deviations from given contexts and therefore entails higher $coherent_\%$
	\item \textit{More diverse generation allows for more diverse exercise contexts} \\
	A higher value for the $generation_{nucleus}$ instructs the NLM to generate more creative text. This allows the model to generate coherent text for more diverse exercise contexts and entails higher $content_H$
	\item \textit{More strict coherence requirements entail more coherent constituents} \\
	A lower (higher) threshold for detecting incoherencies via $coherence_{cosdist}$ ($coherence_{NSP}$) leads to more exercises that are rated as coherent and therefore entails higher $coherent_\%$
	\item \textit{More strict coherence requirements entail more waiting time} \\
	A lower (higher) threshold for detecting incoherencies via $coherence_{cosdist}$ ($coherence_{NSP}$) leads to more exercises that are labelled as incoherent by the system, entailing lower $success_\%$
	\item \textit{Less strict coherence requirements allow for more diverse exercise contexts} \\
	A higher (lower) threshold for detecting incoherencies via $coherence_{cosdist}$ ($coherence_{NSP}$) leads to less exercises that are labelled as invalid by the system, allowing for more diverse exercise contexts at the cost of coherence. This entails higher $content_H$ and lower $coherence_\%$
	\item \textit{Harder exercises allow for more diverse constituent structures} \\
	Harder exercises (higher $generation_{hardness}$) lead to more complex constituent structures, as this is a design choice of the generation controller for the POC
	\item \textit{Harder exercises entail more waiting time} \\
	Harder exercises (higher $generation_{hardness}$) lead to more complex constituent structures with more constituents that are generated in an unconstrained fashion. This in turn leads to more chances for conflicts and incoherencies, entailing lower $success_\%$
\end{enumerate}

Finally, it is observed that $valid_\%$ is not significantly affected by any parameter except exercise hardness. This is due to, as will be shown in the ablation study, that any parameter value for $conflict_{condition}$ has been sufficiently effective to identify invalid exercises. Therefore, there was no clear distinction between its individual parameter values. As the ablation study will show, having no conflict checking included at all entails significantly lower $valid_\%$.

\begin{table}[]
	\centering
		\begin{tabular}{|l|c|c|c|c|c|c|}
			\hline
			\rowcolor[HTML]{000000} 
			{\color[HTML]{FFFFFF} \textbf{Correlation}} &
			\multicolumn{1}{c|}{\cellcolor[HTML]{000000}{\color[HTML]{FFFFFF} $success_\%$}} &
			\multicolumn{1}{c|}{\cellcolor[HTML]{000000}{\color[HTML]{FFFFFF} $coherent_\%$}} &
			{\color[HTML]{FFFFFF} $valid_\%$} &
			{\color[HTML]{FFFFFF} $content_H$} &
			{\color[HTML]{FFFFFF} $arrangement_H$} &
			{\color[HTML]{FFFFFF} abs. Effect} \\ \hline
			\cellcolor[HTML]{000000}{\color[HTML]{FFFFFF} $generation_{hardness}$} &
			\cellcolor{myred!30}-0.33 &
			0.22 &
			\cellcolor{myred!30}-0.31 &
			-0.18 &
			\cellcolor{myblue}0.78 &
			\textbf{1.81} \\ \hline
			\cellcolor[HTML]{000000}{\color[HTML]{FFFFFF} $generation_{nucleus}$} &
			\cellcolor{myred!45}-0.37 &
			\cellcolor{myred}-0.56 &
			0.14 &
			\cellcolor{myblue!65}0.49 &
			-0.10 &
			\textbf{1.66} \\ \hline
			\cellcolor[HTML]{000000}{\color[HTML]{FFFFFF} $coherence_{cosdist}$} &
			\cellcolor{myblue!60}0.46 &
			-0.21 &
			-0.19 &
			\cellcolor{myblue!60}0.47 &
			0.23 &
			\textbf{1.57} \\ \hline
			\cellcolor[HTML]{000000}{\color[HTML]{FFFFFF} $coherence_{NSP}$} &
			\cellcolor{myred!80}-0.45 &
			\cellcolor{myblue!45}0.37 &
			0.01 &
			\cellcolor{myred!30}-0.34 &
			-0.20 &
			\textbf{1.37} \\ \hline
			\cellcolor[HTML]{000000}{\color[HTML]{FFFFFF} $conflict_{condition}$} &
			0.29 &
			-0.25 &
			-0.06 &
			0.30 &
			-0.06 &
			\textbf{0.96} \\ \hline
		\end{tabular}%
	\caption{Measured correlation of $(parameter,response)$ pairs (own illustration)}
	\label{tab:corr}
\end{table}

\begin{table}[]
	\centering
		\begin{tabular}{|
				>{\columncolor[HTML]{000000}}l |c|c|c|c|c|c|}
			\hline
			{\color[HTML]{FFFFFF} \textbf{Observations}} &
			\cellcolor[HTML]{000000}{\color[HTML]{FFFFFF} $success_\%$} &
			\cellcolor[HTML]{000000}{\color[HTML]{FFFFFF} $coherent_\%$} &
			\multicolumn{1}{l|}{\cellcolor[HTML]{000000}{\color[HTML]{FFFFFF} $valid_\%$}} &
			\multicolumn{1}{l|}{\cellcolor[HTML]{000000}{\color[HTML]{FFFFFF} $content_H$}} &
			\multicolumn{1}{l|}{\cellcolor[HTML]{000000}{\color[HTML]{FFFFFF} $arrangement_H$}} &
			\multicolumn{1}{l|}{\cellcolor[HTML]{000000}{\color[HTML]{FFFFFF} abs. Effect}} \\ \hline
			{\color[HTML]{FFFFFF} $generation_{hardness}$}            & 8 &   & 8 &   & 7 &\\ \hline
			{\color[HTML]{FFFFFF} $generation_{nucleus}$} & 1 & 2 &   & 3 &  & \\ \hline
			{\color[HTML]{FFFFFF} $coherence_{cosdist}$}            & 5 &   &   & 6 & &  \\ \hline
			{\color[HTML]{FFFFFF} $coherence_{NSP}$}            & 5 & 4 &   & 6 &  & \\ \hline
			{\color[HTML]{FFFFFF} $conflict_{condition}$}                &   &   &   &   & &  \\ \hline
		\end{tabular}%
	\caption{Connection of derivations to measurements (own illustration)}
	\label{tab:obs}
\end{table}

\newpage

\subsection{Ablation Study}

The following section presents the results of the ablation study. As outlined above in figure \ref{fig:funnel}, the following configurations of the system are tested:
\begin{enumerate}
	\item \textit{$None$} \\
	Conflict checking and coherence checking are omitted
	\item \textit{$Conflict$} \\
	Conflict checking is performed, coherence checking is omitted
	\item \textit{$Coherence_{cosdist+NSP}$} \\
	Coherence checking is performed with cosine distance and NSP criteria
	\item \textit{$Conflict + Coherence_{cosdist}$} \\
	Conflict checking is performed, coherence checking is performed only with cosine distance criteria
	\item \textit{$Conflict + Coherence_{NSP}$} \\
	Conflict checking is performed, coherence checking is performed only with NSP criteria
	\item \textit{$Full$} \\
	Nothing is omitted
\end{enumerate}

The ablation study is performed with parameters optimized through the insights provided in section 4.2, namely:
\begin{enumerate}
	\item \textit{$generation_{nucleus} = 0.5$}
	\item \textit{$coherence_{cosdist} = 0.3$}
	\item \textit{$coherence_{NSP} = 0.99$}
	\item \textit{$conflict_{condition} = (0.5,0.2)$}
\end{enumerate}

Table \ref{tab:abl_coh} shows the response $coherent_\%$ in regards to different system configurations. It is observed that easy exercises with less complex constituent structures are not affected as heavily by omitting checking procedures, whereas harder exercises show significant degradation of $coherent_\%$. This is in line with expectations, as simple constituents structures can more easily be generated in a first-time-right fashion than complex arrangements, which require thorough quality control. It shall be noted that different parameter configurations than defined above lead to, as one would expect, the highest values for $coherence_\%$ for easy exercises, namely 90\%. This can be achieved by setting $coherence_{cosdist} = 0.2$. This anomaly, where exercise coherence degrades with simpler constituent structure, is to be evaluated in further studies.

Table \ref{tab:abl_val} shows the response of $valid_\%$. A clear dependency of $valid_\%$ on $Coherence_{cosdist}$ is evident. This is in line with table \ref{tab:corr}, where $coherence_{cosdist}$ displays the largest absolute value of correlation to $valid_\%$ excluding exercise hardness. Additionally, a less significant contribution of dedicated conflict checking is evident. Coherence checking and confidence checking therefore cannot be separated completely, though, on further inspection of records, it is evident that $Conflict$ is especially valuable in detecting numerical conflicts. The sentence embeddings employed by $Coherence_{cosdist}$ seem to not detect such details and instead focus on semantic attributes.

Table \ref{tab:abl_suc} shows the response of $success_\%$. Again, a strong dependence on $Coherence_{cosdist}$ is evident. This entails that most exercise instances are labelled as incoherent by the system. On closer inspection it is noted, that this procedure also leads to numerous false positives, i.e. exercise instances being excluded for incoherence even though human ratings find them coherent. This implies a tradeoff for the user of the system between waiting time and exercise validity. Recalling table \ref{tab:abl_coh} and table \ref{tab:abl_val}, one can argue that at least for easy exercises a faster version of the system with only conflict checking can serve as an interesting option for interactive deployments. Since the degradation of $coherent_\%$ is not as drastic for easy exercises as for harder ones, this tradeoff might be worthwile considering.

Finally, table \ref{tab:abl_con} shows the response of $content_H$. As expected, $content_H$ decreases with more checking mechanisms. Interestingly, the complete system without any omissions performs very similar to $Coherence_{cosdist+NSP}$, implying that most of the decrease in $content_H$ can be attributed to both $Coherence_{cosdist}$ and $Coherence_{NSP}$. The observation that $Conflict$ entails only a slight decrease of $content_H$ in comparison to $None$ supports this derivation. Again, coherence checking is at the heart of a tradeoff, in this case between diversity and validity.
Similar observations are made for $arrangement_H$.

\newpage

\begin{tightcenter}
	\begin{table}[ht]
		\centering
		\begin{tabular}{|
				>{\columncolor[HTML]{000000}}l |c|c|c|}
			\hline
			{\color[HTML]{FFFFFF} \textbf{Coherence\textsubscript{\%}}} &
			\cellcolor[HTML]{000000}{\color[HTML]{FFFFFF} $Easy$} &
			\cellcolor[HTML]{000000}{\color[HTML]{FFFFFF} $Medium$} &
			\multicolumn{1}{l|}{\cellcolor[HTML]{000000}{\color[HTML]{FFFFFF} $Hard$}} \\ \hline
			{\color[HTML]{FFFFFF} $None$}      & 71 & \cellcolor{myred!70}57 & \cellcolor{myred}52 \\ \hline
			{\color[HTML]{FFFFFF} $Conflict$}      & 70 & \cellcolor{myred!50}59 & \cellcolor{myred!70}57                         \\ \hline
			{\color[HTML]{FFFFFF} $Coherence_{cosdist + NSP}$}  & 73 & 73 & \cellcolor{myred!40}63                         \\ \hline
			{\color[HTML]{FFFFFF} $Conflict + Coherence_{cosdist}$} & 72 & 70 & \cellcolor{myred!20}66                         \\ \hline
			{\color[HTML]{FFFFFF} $Conflict + Coherence_{NSP}$} & \cellcolor{myblue!20}74 & \cellcolor{myred!20}65 & 73                         \\ \hline
			{\color[HTML]{FFFFFF} $Complete$}      & \cellcolor{myblue!60}79 & \cellcolor{myblue}87 & \cellcolor{myblue!80}84                         \\ \hline
		\end{tabular}%
		\caption{$coherence_\%$ in response to ablation of modules (own illustration)}
		\label{tab:abl_coh}
	\end{table}
	\vspace{-0.5cm}
	\begin{table}[ht]
		\centering
		\begin{tabular}{|
				>{\columncolor[HTML]{000000}}l |c|c|c|}
			\hline
			{\color[HTML]{FFFFFF} \textbf{Valid\textsubscript{\%}}} &
			\cellcolor[HTML]{000000}{\color[HTML]{FFFFFF} $Easy$} &
			\cellcolor[HTML]{000000}{\color[HTML]{FFFFFF} $Medium$} &
			\multicolumn{1}{l|}{\cellcolor[HTML]{000000}{\color[HTML]{FFFFFF} $Hard$}} \\ \hline
			{\color[HTML]{FFFFFF} $None$}      & \cellcolor{myred!50}91 & \cellcolor{myred!80}83 & \cellcolor{myred}82 \\ \hline
			{\color[HTML]{FFFFFF} $Conflict$}      & \cellcolor{myred!35}94 & 97 & \cellcolor{myred!15}96                         \\ \hline
			{\color[HTML]{FFFFFF} $Coherence_{cosdist + NSP}$}  & \cellcolor{myblue}100 & \cellcolor{myblue}100 & \cellcolor{myblue!70}99                         \\ \hline
			{\color[HTML]{FFFFFF} $Conflict + Coherence_{cosdist}$} & \cellcolor{myblue}100 & \cellcolor{myblue!70}99 & \cellcolor{myblue}100                         \\ \hline
			{\color[HTML]{FFFFFF} $Conflict + Coherence_{NSP}$} & 97 & \cellcolor{myred!35}93 & \cellcolor{myred!15}96                          \\ \hline
			{\color[HTML]{FFFFFF} $Complete$}      & \cellcolor{myblue}100 & \cellcolor{myblue!70}99 & \cellcolor{myblue!70}99                         \\ \hline
		\end{tabular}%
		\caption{$valid\%$ in response to ablation of modules (own illustration)}
		\label{tab:abl_val}
	\end{table}
	\vspace{-0.5cm}
	\begin{table}[ht]
		\centering
		\begin{tabular}{|
				>{\columncolor[HTML]{000000}}l |c|c|c|}
			\hline
			{\color[HTML]{FFFFFF} \textbf{Success\textsubscript{\%}}} &
			\cellcolor[HTML]{000000}{\color[HTML]{FFFFFF} $Easy$} &
			\cellcolor[HTML]{000000}{\color[HTML]{FFFFFF} $Medium$} &
			\multicolumn{1}{l|}{\cellcolor[HTML]{000000}{\color[HTML]{FFFFFF} $Hard$}} \\ \hline
			{\color[HTML]{FFFFFF} $None$}      & \cellcolor{myblue}100 & \cellcolor{myblue}100 & \cellcolor{myblue}100 \\ \hline
			{\color[HTML]{FFFFFF} $Conflict$}      & \cellcolor{myblue!70}86 & \cellcolor{myblue!55}74 & \cellcolor{myblue!55}76                         \\ \hline
			{\color[HTML]{FFFFFF} $Coherence_{cosdist + NSP}$}  & \cellcolor{myred!90}7 & \cellcolor{myred}3 & \cellcolor{myred}3                       \\ \hline
			{\color[HTML]{FFFFFF} $Conflict + Coherence_{cosdist}$} & \cellcolor{myblue!60}11 & \cellcolor{myred!90}8 & \cellcolor{myred!90}6                         \\ \hline
			{\color[HTML]{FFFFFF} $Conflict + Coherence_{NSP}$} & \cellcolor{myblue!30}56 & \cellcolor{myblue!20}42 & \cellcolor{myblue!15}36                          \\ \hline
			{\color[HTML]{FFFFFF} $Complete$}      & \cellcolor{myred!90}6 & \cellcolor{myred}2 & \cellcolor{myred}2                         \\ \hline
		\end{tabular}%
		\caption{$success\%$ in response to ablation of modules (own illustration)}
		\label{tab:abl_suc}
	\end{table}
	\vspace{-0.5cm}
	\begin{table}[!ht]
		\centering
		\begin{tabular}{|
				>{\columncolor[HTML]{000000}}l |c|c|c|}
			\hline
			{\color[HTML]{FFFFFF} \textbf{Content\textsubscript{H}}} &
			\cellcolor[HTML]{000000}{\color[HTML]{FFFFFF} $Easy$} &
			\cellcolor[HTML]{000000}{\color[HTML]{FFFFFF} $Medium$} &
			\multicolumn{1}{l|}{\cellcolor[HTML]{000000}{\color[HTML]{FFFFFF} $Hard$}} \\ \hline
			{\color[HTML]{FFFFFF} $None$}      & \cellcolor{myblue}5.40 & \cellcolor{myblue}5.40 & \cellcolor{myblue}5.40 \\ \hline
			{\color[HTML]{FFFFFF} $Conflict$}      & \cellcolor{myblue!60}5.35 & \cellcolor{myblue!55}5.32 & \cellcolor{myblue!60}5.33                         \\ \hline
			{\color[HTML]{FFFFFF} $Coherence_{cosdist + NSP}$}  & \cellcolor{myred!60}4.78 & \cellcolor{myred}4.50 & \cellcolor{myred!60}4.80                       \\ \hline
			{\color[HTML]{FFFFFF} $Conflict + Coherence_{cosdist}$} & 4.95 & 4.93 & 4.93                        \\ \hline
			{\color[HTML]{FFFFFF} $Conflict + Coherence_{NSP}$} & \cellcolor{myblue!30}5.15 & \cellcolor{myblue!20}5.11 & 4.91                          \\ \hline
			{\color[HTML]{FFFFFF} $Complete$}      & \cellcolor{myred!75}4.70 & \cellcolor{myred!60}4.80 & \cellcolor{myred!80}4.66                         \\ \hline
		\end{tabular}%
		\caption{$content_H$ in response to ablation of modules (own illustration)}
		\label{tab:abl_con}
	\end{table}
	\vspace{-0.5cm}
\end{tightcenter}

\section{Discussion}

The results presented in section 4 testify the ability of the proposed approach to be effective in generating valid, coherent and diverse training exercises in the field of mathematical statistics. A tradeoff between generation time and exercise validity is evident, which can be chosen flexibly according to specific needs. In an environment where teachers are readily available to resolve any conflicts manually and exercise throughput is key, generation speed can be favored. On the other hand, in an enviroment implies full automation without the possibility of human intervention, validity can be prioritized.

One positive aspect of the approach proposed in section 3 lies within its modularity, allowing for individual scalability of submodules and clear attribution of specific subtasks. Though, in the course of the development of the POC, specific tailoring of the interaction between the submodules has proven to be a vital aspect for the system as a whole. An example of this is the interaction of the submodule for constraint checking and constraint generation (recall figure \ref{fig:architecture-approach}). In order to increase the ability of the constraint checking module, the design of generation templates needs to be tailored to contain certain keywords such as 'only' or 'exclusively' so that conflicts become more pronounced.

In a similar fashion, the user benefits from being aware about these interactions when planning their inputs, i.e. the exercise prompt and exercise definition. For example, by setting up the prompt in a way that allows smooth transitioning to the automatically generated constraints the user can generate valid exercises much quicker than otherwise. Though helpful, no specific knowledge of the inner workings of NLMs is required for this. The nature of these precautions includes formulating the prompt in an open ended fashion and containing enough context information in it for the unconstrained generation module to follow up on. This skill can be acquired through generating a handful of exercises. 

Finally, in addition to the desire of being able to generate diverse training exercises in order to allow students for abstraction of underlying concepts instead of shallow surface structure, the author notes that another perspective on supporting the learning experience has been noticed in the course of testing the system. The mere act of thinking ahead for the system in terms of planning how both the prompt and exercise configuration are provided to the system has proven to be a valuable change of perspective for understanding the concepts at hand. This design choice entails the (coincidental) positive aspect of forcing the student in the position of a teacher, leading to an understanding of exercises from a completely different perspectives. The detailed thought process of questions such as "What even is the random variable at hand?" and "How does my variable fit into the provided context?" might lead to further boosts in the abstraction of concepts of interest.

\section{Conclusion}
This thesis proposes a novel approach to synthesize natural language subject to general content validity constraints. This approach is applied to generating word problems for mathematical statistics and is proven to be effective in terms of constraint satisfaction and diversity. 
In contrast to related work that focuses largely on monolithic approaches, the proposed method harnesses the benefits of a modular design, namely independent scalability of subtasks and straight forward extension of scope without complete retraining.
The experimental results present a tradeoff between generation time and exercise validity. The system can easily be parametrized to handle this tradeoff according to the requirements of specific use cases.
For future work, it is worth exploring ways to extend the scope of applicability of the proposed system by including additional functionality, striving for an integrated solution to facilitate the complete process of exercise design, exercise processing and solution checking.

\printbibliography






\end{document}